\documentclass[runningheads]{llncs}

\usepackage{eccv}

\usepackage{eccvabbrv}

\usepackage[accsupp]{axessibility}  

\usepackage{multirow}
\usepackage{array}
\usepackage{makecell}   
\usepackage{graphicx}
\usepackage{amsfonts}   
\usepackage{amsmath}    
\usepackage{booktabs}   
\usepackage[table,xcdraw]{xcolor}   
\usepackage{bm}         
\usepackage{color}      
\usepackage{algorithm}
\usepackage{algorithmic}
\usepackage{setspace}

\usepackage{tikz}
\usepackage{pgfplots}
\usepackage{amssymb}
\usepackage{adjustbox}

\usepackage{pifont}
\definecolor{lightblue}{RGB}{230, 245, 255}
\usepackage{dsfont}
\usepackage{wrapfig}

\usepackage{hyperref}

\usepackage{orcidlink}

\begin{document}

\title{UniSem: Generalizable Semantic 3D Reconstruction from Sparse Unposed Images}

\titlerunning{UniSem}

\author{Guibiao Liao \and
Qian Ren \and 
Kaimin Liao \and 
Hua Wang \and 
Zhi Chen \and 
Luchao Wang \and
Yaohua Tang
}

\authorrunning{G. Liao et al.}

\institute{Moore Threads AI \\
}
\vspace{-10mm}

\maketitle

\begin{abstract}

    Semantic-aware 3D reconstruction from sparse, unposed images remains challenging for feed-forward 3D Gaussian Splatting (3DGS). 
    Existing methods often predict an over-complete set of Gaussian primitives under sparse-view supervision, leading to unstable geometry and inferior depth quality. 
    Meanwhile, they rely solely on 2D segmenter features for semantic lifting, which provides weak 3D-level and limited generalizable supervision, resulting in incomplete 3D semantics in novel scenes.
    To address these issues, we propose UniSem, a unified framework that jointly improves depth accuracy and semantic generalization via two key components. 
    First, Error-aware Gaussian Dropout (EGD) performs error-guided capacity control by suppressing redundancy-prone Gaussians using rendering error cues, producing meaningful, geometrically stable Gaussian representations for improved depth estimation. 
    Second, we introduce a Mix-training Curriculum (MTC) that progressively blends 2D segmenter-lifted semantics with the model’s own emergent 3D semantic priors, implemented with object-level prototype alignment to enhance semantic coherence and completeness.
    Extensive experiments on ScanNet and Replica show that UniSem achieves superior performance in depth prediction and open-vocabulary 3D segmentation across varying numbers of input views. 
    Notably, with 16-view inputs, UniSem reduces depth Rel by 15.2\% and improves open-vocabulary segmentation mAcc by 3.7\% over strong baselines. 
    
\end{abstract}


\begin{figure*}
\centering
\includegraphics[width=\linewidth]{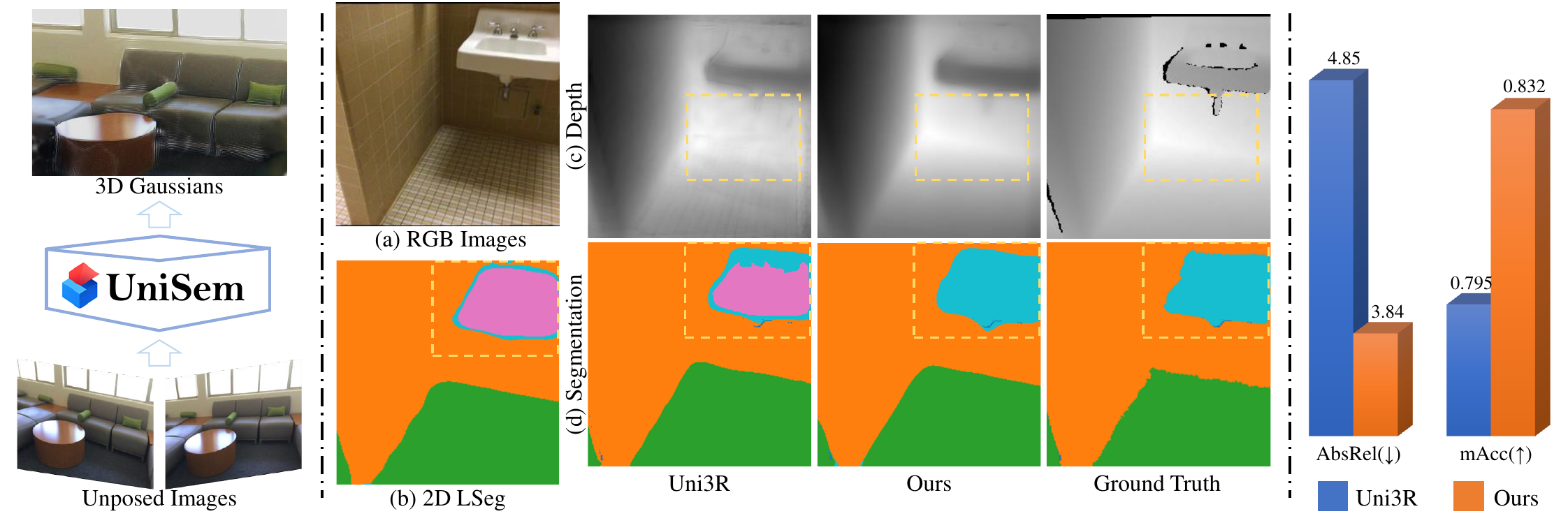}
\vspace{-6mm}
\captionof{figure}{
UniSem, a feedforward Gaussian model for 3D semantic-aware reconstruction, predicts a unified 3D Gaussian scene representation from unposed images in a single forward pass, enabling novel view synthesis, depth estimation, and open-vocabulary segmentation. 
Prior work Uni3R \cite{uni3r} predicts pixel-aligned Gaussians and relies solely on 2D LSeg features \cite{LSeg} for semantic lifting, resulting in \textit{noisy depth} and \textit{limited semantic generalization (golden dashed box)}. 
In contrast, UniSem suppresses redundant Gaussians and complements 2D lifting with emergent 3D semantic cues, yielding more accurate depth and more coherent, complete 3D semantics. 
} 
\label{fig:vis_fig1}
\vspace{-6mm}
\end{figure*}

\section{Introduction} \label{sec:intro} 
Jointly reconstructing and understanding 3D scenes from 2D multi-view images (\ie, semantic-aware 3D reconstruction) is a long-standing challenge in computer vision.
Recent advances in radiance fields, most notably Neural Radiance Fields (NeRFs) \cite{nerf} and 3D Gaussian Splatting (3DGS) \cite{3dgs}, have substantially improved semantic-aware 3D reconstruction. 
A prominent line of work distills 2D vision-language foundation model features (\eg, CLIP \cite{OpenAICLIP, OpenCLIP}) into radiance fields, enabling open-vocabulary 3D understanding \cite{LERF,3DOVS,LEGaussian,langsplat,gaugrouping}. 
However, these methods typically rely on multi-stage preprocessing (\eg, SfM-based pose estimation \cite{sfm,colmap}) and per-scene optimization, which restricts scalability and the ability to understand novel scenes.

Recent efforts \cite{splatt3r,lvsm,anysplat} have shifted toward generalizable feed-forward reconstruction, enabling single-pass inference without scene-specific optimization and improving cross-scene view synthesis.  
Early methods \cite{pixelsplat,mvsplat,freesplat,monosplat} predict pixel-aligned Gaussians from known posed images. 
Recent works \cite{dust3r,fast3r,noposplat,vggt} relax the pose requirement, enabling generalizable reconstruction under unposed settings. 
More recently, LSM \cite{lsm} and Uni3R \cite{uni3r} incorporate 2D segmenter guidance to advance semantic 3D reconstruction from sparse, unposed images.

Despite these advances, existing generalizable semantic 3DGS methods still face two challenges that undermine a \emph{geometrically stable} and \emph{semantically coherent} 3D representation: 
\textbf{(1) Over-complete pixel-granular Gaussians introduce depth noise.}
A common feed-forward 3DGS pipeline predicts pixel-aligned Gaussians (\ie, one Gaussian per pixel), yielding an over-complete primitive set under sparse inputs.
Redundant primitives compete to explain the same photometric evidence, causing gradient interference and resulting in noisy depth estimates (Fig.~\ref{fig:vis_fig1}(c)).
\textbf{(2) 2D-only lifting limits semantic generalization.} 
Purely lifting view-local 2D segmenter features provides weak 3D-consistent supervision and is bounded by the generalization ceiling of the 2D teacher, limiting the coherence and completeness of 3D semantics in novel scenes (Fig.~\ref{fig:vis_fig1}(d)).

To address these issues, we propose \textbf{UniSem}, a unified framework that incorporates \textbf{Error-aware Gaussian Dropout (EGD)} and \textbf{Mix-training Curriculum (MTC)} to jointly enhance depth quality and semantic accuracy.

First, in reliably reconstructed (low-error) regions, excessively accumulated Gaussians tend to be redundant. They contribute marginally to rendering fidelity and sometimes introduce noisy gradients that harm depth estimation.
To mitigate this, EGD assigns each Gaussian an error-aware dropout probability during training, suppressing gradient noise from redundancy-prone primitives. 
This simple yet effective regularization encourages a more compact and stable Gaussian distribution, yielding improved depth accuracy (Fig. \ref{fig:vis_fig1} (c)).

Second, as a feed-forward 3D model gradually acquires cross-scene knowledge, it develops \textit{emergent 3D semantic cues} that can be more coherent and generalizable than raw 2D-lifted features. 
Motivated by this, MTC injects such 3D and generalizable cues into the supervision loop: training starts with purely 2D-lifted semantics, after which model-predicted transferable semantics are blended to progressively guide learning. 
This yields more coherent and complete 3D semantics across diverse unseen scenes (Fig. \ref{fig:vis_fig1} (d)).

Together, EGD improves geometric stability, and MTC enhances semantic generalization, jointly refining Gaussian representations for more accurate semantic-aware 3D reconstruction. 
Our contributions are summarized as follows:
\vspace{-6.5mm}
\begin{itemize}
    \item We propose UniSem, an effective generalizable semantic 3D reconstruction model that improves depth quality and semantic accuracy. 
    \item We introduce Error-aware Gaussian Dropout (EGD), a lightweight yet effective capacity control mechanism that suppresses redundancy-induced Gaussians and boosts depth prediction. 
    \item We present Mix-training Curriculum (MTC) to integrate emergent 3D semantic priors, improving semantic coherence and cross-scene generalization.
    \item UniSem achieves superior performance on depth prediction and open-vocabulary 3D segmentation across diverse scenes and view input settings, showing strong generalization and scalability. 
\end{itemize}

\section{Related Work} \label{sec:related}
\textbf{Per-Scene Optimization-based 3D Reconstruction.} 
Radiance-field representations have rapidly advanced 3D reconstruction and novel-view synthesis in recent years.
Neural Radiance Fields (NeRFs) \cite{nerf, nerfreview} implicitly model scenes via coordinate-based MLPs and achieve photo-realistic rendering, while 3D Gaussian Splatting (3DGS) \cite{3dgs, wu2024recent} represents scenes with explicit Gaussian primitives and enables efficient rasterization with high fidelity. 
Despite their success, most NeRF/3DGS pipelines \cite{zipnerf,instantngp,scaffoldgs,mipsplatting,2dgs}, rely on multi-stage preprocessing (\eg, SfM for camera poses/Gaussian initialization) and require costly per-scene optimization.
Such workflows limit scalability and hinder cross-scene generalization.

\noindent
\textbf{Generalizable Feed-forward Reconstruction.}
To eliminate per-scene optimization, generalizable feed-forward models have emerged.
Early, pixelNeRF \cite{pixelnerf} and MVSNeRF \cite{mvsnerf} learn neural priors from large-scale datasets, enabling unseen scene reconstruction. 
pixelSplat \cite{pixelsplat}, MVSplat \cite{mvsplat}, and subsequent works \cite{depthsplat, nam2025generative} directly predict Gaussian primitives for efficient feed-forward rendering. 
Yet, these methods still depend on pre-computed camera poses.

Recently, pose-free methods relax this assumption. DUSt3R \cite{dust3r} and MASt3R \cite{mast3r} infer pixel-aligned 3D point clouds from unposed image pairs, and VGGT \cite{vggt} jointly predicts structure and poses using stacked Transformer blocks.
These advances improve feed-forward geometry and appearance modeling, but most do not explicitly address open-vocabulary segmentation. 
Instead, our goal is a unified model capable of generalizable view synthesis and semantic reconstruction.

\noindent
\textbf{Generalizable Semantic 3D Reconstruction.}
Vision-language models such as CLIP \cite{OpenAICLIP,OpenCLIP} have inspired efforts to lift 2D open-vocabulary knowledge into radiance fields.
LERF \cite{LERF} distills CLIP features into NeRF, and 3DGS-based extensions \cite{langsplat, gaugrouping, feature3dgs} augment Gaussians with semantic embeddings for efficient 3D understanding. 
However, these methods typically follow per-scene optimization pipelines, limiting scalability to novel scenes.

Recent works target feed-forward semantic reconstruction without per-scene optimization.
LSM \cite{lsm} and GSemSplat \cite{gsemsplat} use Transformers to encode cross-scene semantic priors, while UniForward \cite{uniforward} decouples appearance/semantics with a dual-branch decoder. 
SpatialSplat \cite{spatialsplat} designs a dual-field semantic representation to learn uncompressed semantics.  
SIU3R~\cite{siu3r} and IGGT~\cite{IGGT} further bridge reconstruction and instance-level understanding, but both rely on instance annotations.
However, most of these methods are restricted to two-view inputs. 
More recently, Uni3R~\cite{uni3r} introduces cross-frame attention over unposed multi-view inputs, enabling more flexible multi-view semantic reconstruction.

Despite these advances, over-complete pixel-aligned Gaussians and 2D-only semantic lifting limit depth quality and cross-scene semantic generalization.
Our approach addresses them via error-guided Gaussian redundancy suppression and curriculum-weighted mixed semantic supervision.

\section{Empirical Observations}\label{sec:observations}
\noindent\textit{\textbf{Redundant Gaussians are photometrically negligible yet geometrically harmful.}} 
Feed-forward 3DGS pipelines commonly predict pixel-aligned Gaussians, which can become over-complete under sparse inputs. 
To understand the 
\begin{wrapfigure}{r}{0.51\linewidth}
\vspace{-25pt}
\centering
\includegraphics[width=0.98\linewidth]{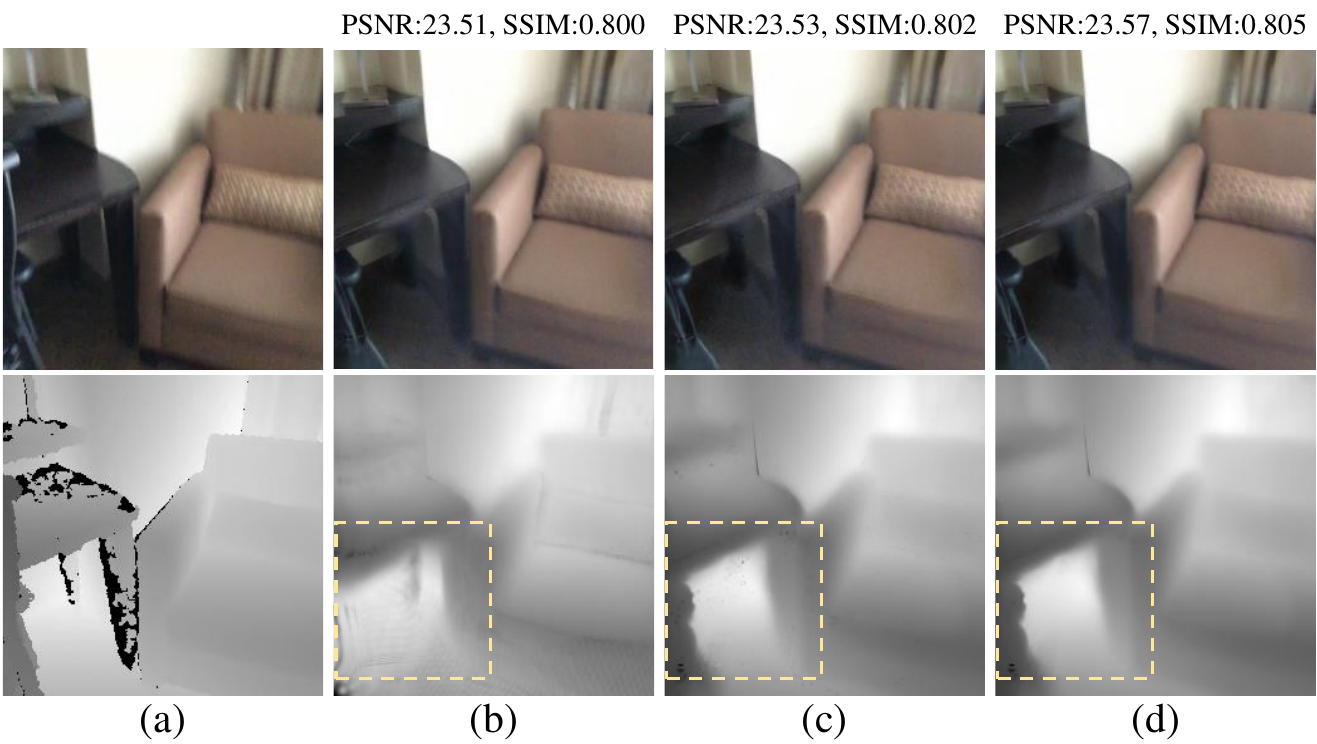}
\vspace{-2mm}
\caption{
Visual results under different settings in feed-forward 3DGS. 
(a) Ground truth.
(b) Baseline: redundant Gaussians persist, leading to noisy depth.
(c) Removal of a subset of Gaussians in low-error regions (with short re-optimization): appearance changes are minimal while depth noise is reduced. 
(d) After continued training with redundancy suppression, the model yields more stable and accurate depth. 
}
\label{fig:vis_analysis_egd}
\vspace{-25pt}
\end{wrapfigure}
role of redundancy, we conduct a diagnostic experiment during training by removing a subset of Gaussians in reliably reconstructed (low-error) regions according to the per-pixel reconstruction error. 
As shown in Fig.~\ref{fig:vis_analysis_egd}(c), after removing a subset of Gaussians in low-error regions and re-optimizing for a short period (1k iterations), the rendered appearance is preserved, while depth predictions change substantially and become more stable. 
This suggests that many such primitives have limited photometric impact but can introduce gradient interference that harms geometric learning.
With continued training under redundancy suppression, the model further yields more stable and accurate depth (Fig.~\ref{fig:vis_analysis_egd}(d)).

\noindent\textbf{Takeaway.}
These results suggest that suppressing redundancy-prone Gaussians, especially in low-error regions, can preserve appearance while improving depth stability, motivating an error-guided mechanism to control effective primitive capacity during training.

\begin{figure}[t]
\centering
\begin{minipage}[t]{0.49\textwidth}
  \vspace{0pt}\centering
  \includegraphics[width=.98\linewidth]{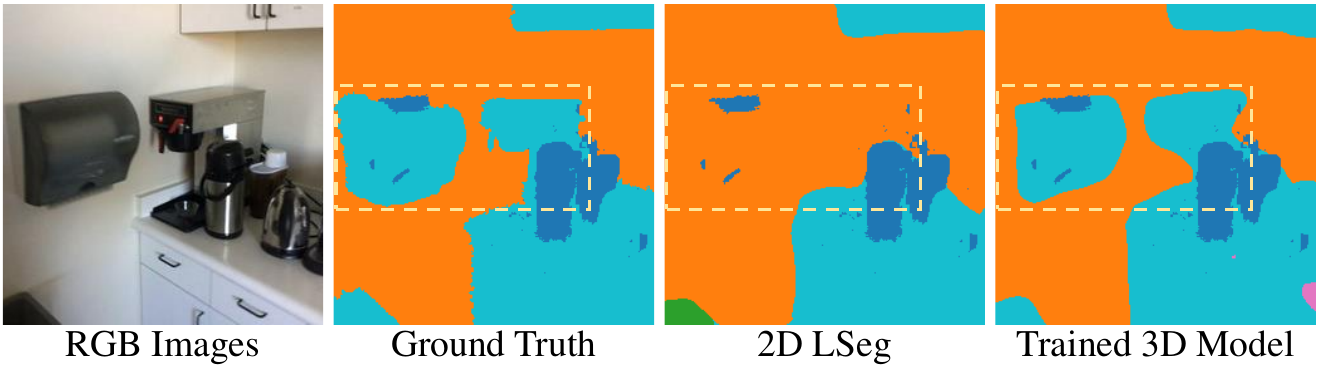}
  \vspace{-2.5mm}
  \captionof{figure}{ Visual segmentation results of different models. }
  \label{fig:vis_analysis_mtc}
\end{minipage}\hfill%
\begin{minipage}[t]{0.49\textwidth}
  \vspace{0pt}\centering
  \small
  \setlength{\tabcolsep}{3pt}
    \begin{adjustbox} {width=\linewidth}
    \begin{tabular}{ c | c c | c c | c c }
    \Xhline{3\arrayrulewidth} 
    \multirow{2}{*}{Method} & \multicolumn{2}{c|}{2 Views} & \multicolumn{2}{c|}{8 Views} & \multicolumn{2}{c}{16 Views} \\ 
    & mIoU  & mAcc  & mIoU  & mAcc  & mIoU  & mAcc    \\   
    \midrule[0.1pt]
    2D LSeg \cite{LSeg}    & 0.482   & 0.793   & 0.448   & 0.773    & 0.464   & 0.788   \\ 
    Trained 3D Model       & 0.540   & 0.818   & 0.507   & 0.805    & 0.500   & 0.794   \\
    \Xhline{3\arrayrulewidth}
    \end{tabular}
    \end{adjustbox}
    \vspace{-3mm}
    \captionof{table}{Open-vocabulary segmentation: 2D LSeg vs. a feed-forward 3D model trained only with 2D-lifted semantics.}
    \label{table:analysis_mtc}
\end{minipage}
\vspace{-4mm}
\end{figure}

\noindent\textit{\textbf{Feed-forward 3D models can develop generalizable 3D semantic cues beyond 2D lifting.}}
Most semantic 3DGS methods typically lift 2D segmenter features into 3D for supervision, which is view-conditioned and is bounded by the cross-scene generalization ceiling of the 2D teacher. 
Interestingly, we find that a feed-forward 3DGS model trained across scenes can produce semantic predictions that are often \emph{more coherent and complete} than raw 2D segmenter outputs on unseen scenes.
As illustrated in Fig.~\ref{fig:vis_analysis_mtc}, the 2D segmenter fails to recover a complete object region in a novel scene, while the trained 3D model yields a more accurate segmentation.
Table~\ref{table:analysis_mtc} further compares open-vocabulary segmentation metrics between the 2D teacher (LSeg) and the trained feed-forward 3D model supervised only by 2D-lifted semantics.
Across different input-view settings, the trained 3D model can outperform the 2D teacher, supporting the existence of transferable semantic cues emerging in the 3D model beyond raw 2D predictions.

\noindent\textbf{Takeaway.}
These observations motivate leveraging the model's own emergent 3D semantic cues to \emph{complement} 2D-lifted supervision, improving semantic coherence and generalization in novel scenes.

\section{Methodology} \label{sec:method}
Given a sparse set of unposed input images $\{I^v\}_{v=1}^V$, UniSem predicts a 3D Gaussian scene representation in a single forward pass (Sec.~\ref{sec:FeedforwadGS}).
Motivated by the empirical observations in Sec.~\ref{sec:observations}, UniSem incorporates two key components in generalizable semantic 3DGS. 
\textbf{Error-aware Gaussian Dropout (EGD)} for error-guided capacity control to stabilize geometry (Sec.~\ref{sec:EGD}), and a \textbf{Mix-training Curriculum (MTC)} that injects emergent 3D semantic priors to support robust generalization to unseen scenes (Sec.~\ref{sec:MTC}).
Finally, the overall training loss of UniSem is described in Sec. \ref{sec:training}.

\subsection{Feed-Forward Semantic Gaussian Splatting}  \label{sec:FeedforwadGS}
\begin{figure*}[!t]
\centering
\includegraphics[width=\linewidth]{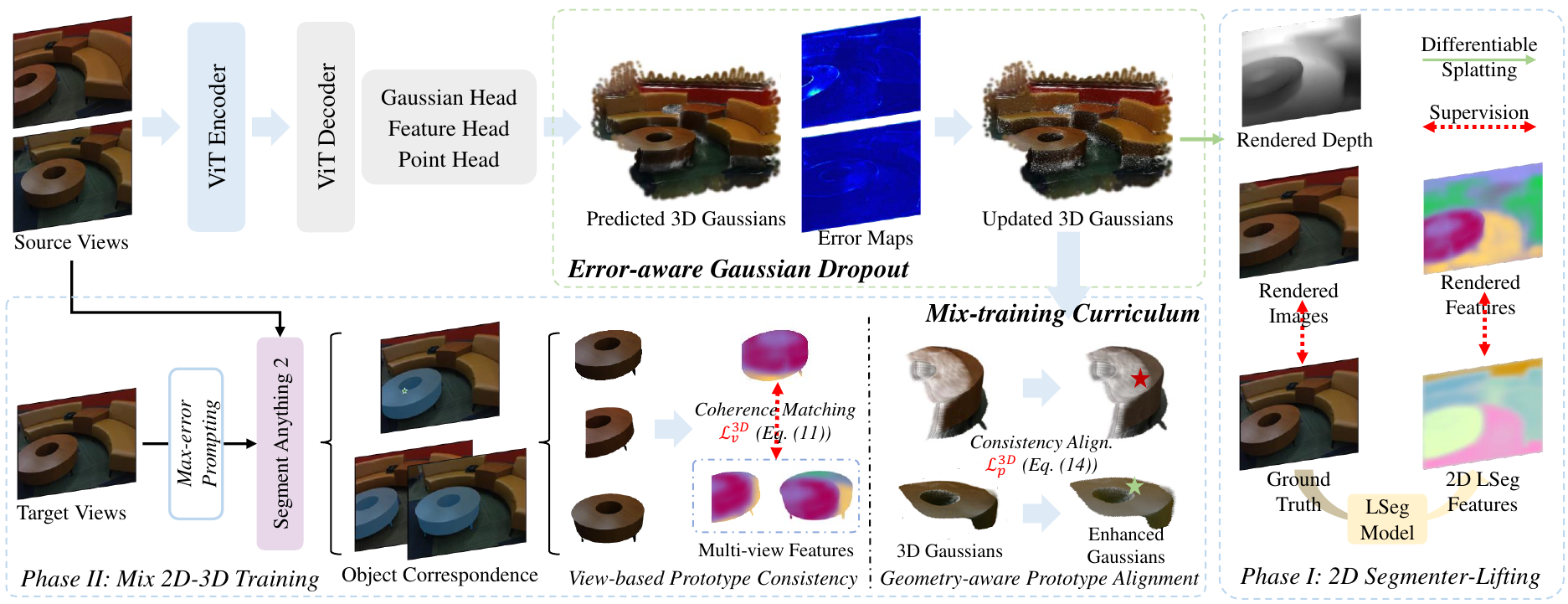}
\vspace{-7mm}
\caption{
\textbf{Overview of UniSem.}
Given a set of unposed input views, a ViT-based encoder extracts multi-view features, which are fed into the decoder and DPT heads to produce pixel-aligned 3D Gaussian primitives. 
(a) To stabilize depth estimation, we introduce \textbf{Error-aware Gaussian Dropout (EGD)}, which computes per-pixel reconstruction error and suppresses overly dense Gaussians via progressive, error-guided dropout, reducing gradient noise during training. 
(b) To enhance 3D semantic consistency and cross-scene generalization, the proposed \textbf{Mix-training Curriculum (MTC)} transitions from pure 2D segmenter-lifted supervision to mixed 2D-3D semantic training. 
MTC leverages (i) max-error prompting for cross-view object correspondence, and (ii) view-to-view and geometry-aware prototype alignment, enforcing more coherent semantic Gaussian representations. 
}
\label{fig:architecture}
\vspace{-4mm}
\end{figure*}

As shown in Fig. \ref{fig:architecture}, UniSem utilizes a Vision Transformer (ViT) \cite{vit} backbone for feature extraction and Dense Prediction Transformer (DPT) \cite{dpt} heads for Gaussian parameter prediction. 

\noindent
\textbf{Transformer Encoder and Feature Fusion.}   
Input images are divided into patch tokens, and are concatenated with a learnable camera intrinsic token \cite{noposplat}. These tokens are processed by a ViT backbone with a DINOv2 architecture \cite{dinov2} for feature extraction. 
Then, these features are processed by a cross-view Transformer encoder (similar to VGG-T \cite{vggt}) that alternates between intra-view self-attention and cross-view attention, effectively fusing multi-view information into a unified latent representation.

\noindent
\textbf{Decoding Gaussian Parameters.} 
Using DPT heads, the fused latent representation is decoded into a dense set of pixel-aligned 3D Gaussian primitives $G_k$ with corresponding parameters. Each Gaussian is parameterized by: 
\begin{equation}
G_{k} = \{\mu_{k}, s_{k}, r_{k}, \alpha_{k}, c_{k}, f_{k}\}, 
\label{eq:gaussian_params}
\end{equation}
where $\mu_{k}$, $s_{k}$, and $r_{k}$ denote the 3D center, scale vector, and rotation quaternion, respectively. $\alpha_{k} \in [0, 1]$ is the opacity. $c_{k}$ is the spherical harmonics (SH) for color representation, and $f_{k} \in \mathbb{R}^d$ is the semantic feature vector. 
More details are provided in the supplementary material.

While this feed-forward pipeline provides an initial semantic 3D Gaussian scene, it often suffers from depth noise due to redundant primitives and limited semantic generalization under 2D-only lifting (Sec.~\ref{sec:observations}), motivating the two components described next.

\subsection{Error-aware Gaussian Dropout (EGD)}  \label{sec:EGD}
We introduce {Error-aware Gaussian Dropout (EGD)} to perform \emph{error-guided capacity control} during training, selectively suppressing redundancy-prone primitives for enhancing depth quality.

\noindent
\textbf{Error-aware Dropout Probability.}
For each pixel-aligned Gaussian $G_k$, we compute the reconstruction error at its associated pixel location $k$: 
\begin{equation}
    E_k = \left\| \hat{I}_k - I_k \right\|_1 ,
\end{equation}
where $\hat{I}_k$ is the rendered pixel obtained via standard $\alpha$-blending of all Gaussians overlapping pixel $k$: $\hat{I}_k = \sum_{i \in \mathcal{N}} {c}_i \alpha_i \prod_{j=1}^{i-1} (1 - \alpha_j)$. 
Then, we normalize the reconstruction error across all pixel-aligned Gaussians: 
\begin{equation}
    \tilde{E}_k = 
    \frac{E_k - \min_j E_j}{\max_j E_j - \min_j E_j + \epsilon}, 
    \label{eq:egd_error_norm}
\end{equation}
with $\epsilon = 10^{-6}$ for numerical stability. 
We assign each Gaussian a dropout probability that decreases with its normalized reconstruction error:
\begin{equation}
    p_k^{\text{drop}} = {\eta}_t \left( 1 - \tilde{E}_k \right)^{\beta},
    \label{eq:egd_probability}
\end{equation}
where $\beta>1$ modulates the focus intensity of low-error regions and $\eta_t$ controls the overall dropout ratio at training epoch $t$.

\noindent
\textbf{Cosine-cycle Schedule.}
To avoid unstable dropout early in training and excessive disturbance in later stages, we modulate $\eta_t$ with a cosine-cycle schedule: 
\begin{equation}
\eta_t = \eta_{\min} + \frac{1}{2}(\eta_{\max} - \eta_{\min})
\left[1 - \cos\!\left(2\pi \frac{t}{T}\right)\right],
\label{eq:cosine_cycle}
\end{equation}
where $t$ and $T$ denote the current and total training epochs. 
We set $\eta_{\max}=0.2$ and $\eta_{\min}=0.05$ in experiments. 

\noindent
\textbf{Stochastic Masking.}
We sample a Bernoulli mask to decide whether $G_k$ participates in rendering and gradient back-propagation:
\begin{equation}
    m_k = \mathds{1}\!\left[\xi_k > p_k^{\text{drop}}\right],
    \qquad 
    \xi_k \sim \mathcal{U}(0,1),
\end{equation}
where $m_k=1$ indicates that Gaussian $G_k$ is retained at this iteration.

EGD is applied \emph{only during training} as a stochastic regularizer that reshapes gradient flow by suppressing redundancy-prone Gaussians. 
At inference, we render with all predicted Gaussians, since EGD training yields a well-conditioned Gaussian set, thus inference-time pruning has a marginal effect (see supplementary). 
As shown in Fig.~\ref{fig:vis_comparison_depth}, using EGD leads to improved depth accuracy.

\subsection{Mix-training Curriculum (MTC)}  \label{sec:MTC}
We introduce a Mix-training Curriculum (MTC) that \emph{complements} 2D-lifted supervision with model-emergent 3D semantic priors for mixed supervision.

\noindent
\textbf{Phase I: 2D Segmenter-Lifted Supervision.}
Following \cite{lsm,uni3r}, we distill features from LSeg~\cite{LSeg} as the initial semantic supervision. 
The rendered semantic feature $\hat{{F}}$ is computed by $\alpha$-blending $\mathcal{N}$ Gaussians overlapping the pixel: 
\begin{equation} 
\hat{{F}} = \sum_{i \in \mathcal{N}} {{f}}_i \alpha_i \prod_{j=1}^{i-1} (1 - \alpha_j), 
\label{eq_alphablending_sem}
\end{equation} 
where ${{f}}_i$ denotes the low-dimensional semantic Gaussian embedding for efficient rasterization.  
We apply a lightweight MLP $\mathcal{F}_{MLP}$ to align $\hat{{F}}$ to the LSeg feature map $\tilde{F}$: 
\begin{equation}
\mathcal{L}_{\text{sem}}^{\text{2D}} = \mathcal{L}_{\cos} \big(\mathcal{F}_{MLP}(\hat{{{F}}}), \tilde{F} \big), 
\label{lseg_cos_loss}  
\end{equation} 
where $\mathcal{L}_{\cos}$ denotes the cosine similarity loss. 
While this step lifts 2D semantics into 3D Gaussians, this purely 2D lifting manner inherits inconsistent noise from 2D segmentations and lacks rich generalization priors.

\noindent
\textbf{Phase II: Mixed 2D-3D Supervision }   
To enhance semantic coherence and generalization, we incorporate the model’s own \textit{3D semantic priors} into the supervision loop. 
Concretely, for $n$ sparse source views ($\{I^{\text{s}_i}\}_{i=1}^{n}$) and a target view ($I^{\text{tar}}$, selected from the intermediate frame among the source views, as in \cite{lsm,uni3r}), we construct object-level correspondences and enforce prototype consistency across views and in 3D.

\noindent
\textit{1) Object Correspondence via Max-error Prompting.} 
We treat the target and source views as a short viewpoint-varying video sequence and use SAM2~\cite{sam2} to obtain corresponding object masks. 
Since directly sampling dense prompts is impractical, we propose Max-error Prompting, which selects the pixel $\mathbf{x}_{\text{max}}$ in the target view exhibiting the highest semantic disagreement as the prompt location. The error map for the target view is computed as: 
\begin{equation}
{E}^{\text{tar}}(x) = 1 - \text{cos} \big( \mathcal{F}_{\text{MLP}}(\hat{F}^{\text{tar}}(x)), \tilde{F}^{\text{tar}}(x) \big). 
\end{equation}  
The prompt location for SAM2 is defined as $\mathbf{x}_{\text{max}}^{\text{tar}} = \arg\max_j {E}^{\text{tar}}(j)$. 
SAM2 then returns a target mask and its corresponding masks in the source views:
\begin{equation}
\mathbf{M}^{\text{tar}}, \{\mathbf{M}^{\text{s}_i}\}_{i=1}^{n} = \text{SAM2} (I^{\text{tar}}, \{I^{\text{s}_i}\}_{i=1}^{n}, \mathbf{x}_{\text{max}}^{\text{tar}}), 
\label{eq:sam2_prompt}
\end{equation}
where $\mathbf{M}^{\text{tar}}$ and $\mathbf{M}^{\text{s}_i}$ denote the region masks for target and $i$-th source views.
SAM2 is used solely during training and is not required at inference.

\noindent
\textit{2) View-to-View Prototype Consistency.}  
Given the matched masks, we impose a view-based prototype consistency constraint to inject 3D semantic priors and suppress 2D noise. 
Specifically, we align the mean semantic features computed from the target and $n$ source views within the matched object region: 
\begin{equation}
\mathcal{L}^{\text{3D}}_v =  \mathcal{L}_{\cos} \left( \frac{\sum_{j \in \mathbf{M}^{\text{tar}}} \hat{F}^{\text{tar}}(j)}{|\mathbf{M}^{\text{tar}}|}, \frac{1}{n} \sum_{i=1}^{n}\frac{\sum_{j \in \mathbf{M}^{\text{s}_i}} \hat{F}^{\text{s}_i}(j)}{|\mathbf{M}^{\text{s}_i}|} \right), 
\label{eq:cross_view_cons}
\end{equation} 
This encourages the model to produce view-invariant 3D semantic representations. 

\noindent
\textit{3) Geometry-aware Prototype Alignment.} 
To further enhance the 3D semantic coherence of Gaussian representations, we propose to explicitly enforce feature matching within the 3D domain. 
For Gaussians within a source-view mask region $\mathbf{M}^{\text{s}}$, we normalize their semantic embeddings $\{ {f}_i \}_{i\in \mathbf{M}^{\text{s}}}$ and form a region prototype by averaging: 
\begin{equation}
\mathbf{p} = \frac{1}{|\mathbf{M}^{\text{s}}|}\sum_{i\in \mathbf{M}^{\text{s}}} {f}_i.
\end{equation}
To account for reliability, each Gaussian is weighted by its local geometric density, measured via mean pairwise distance:
\begin{equation}
d_i = \frac{1}{|\mathbf{M}^{\text{s}}|}\sum_{j\in \mathbf{M}^{\text{s}}}\|\mu_i - \mu_j\|_2. 
\end{equation}
Gaussians located in denser and more structurally stable regions thus receive higher emphasis, yielding a more reliable learning signal. The geometry-aware alignment constraint can be formulated as:
\begin{equation}
\mathcal{L}^{\text{3D}}_p = \sum_{i\in \mathbf{M}^{\text{s}}} w_i \, \big\| {f}_i - \mathbf{p} \big\|_2^2, 
\quad w_i = \frac{d_i^{-1}}{\sum_{k\in \mathbf{M}^{\text{s}}} d_k^{-1}}. 
\end{equation}

\noindent
\textit{4) Total Mix-training Semantic Loss.} 
The overall semantic loss is defined as: 
\begin{equation}
\mathcal{L}_{\text{sem}} = \mathcal{L}_{\text{sem}}^{\text{2D}} + \lambda_{\text{m}}(\mathcal{L}^{\text{3D}}_v + \mathcal{L}^{\text{3D}}_p) \cdot \mathds{1}_{t \geq \tau},
\label{eq:total_sem_loss}
\end{equation}
where $\lambda_{\text{m}}=0.5$ and the curriculum switch $\tau=90$ epochs.  
By injecting semantics derived from 3D generalizable models, our model effectively strengthens semantic consistency, completeness, and cross-scene generalization.

\subsection{Model Training}  \label{sec:training}
Our model is trained end-to-end by minimizing a composite loss, which enforces fidelity across appearance, semantics, and geometric structure: 
\begin{equation}
\mathcal{L} = \mathcal{L}_{\text{color}} + \lambda_1\mathcal{L}_{\text{sem}} + \lambda_2 \mathcal{L}_{\text{geo}}, 
\end{equation}
where $\lambda_1, \lambda_2$ are set to 0.02 and 0.005  in all experiments.

\noindent
\textbf{Photometric Loss ($\mathcal{L}_{\text{color}}$).} 
We use a combination of L1 and the LPIPS perceptual loss \cite{lpips} to ensure the rendered view $\hat{I}_{v}$ matches the input image $I_{v}$:  
\begin{equation}
\mathcal{L}_{\text{color}} = \sum_{v=1}^{V} (\| \hat{I}^{v} - I^{v} \|_{1} + \lambda_{\text{LPIPS}}\mathcal{L}_{\text{LPIPS}}(\hat{I}^{v}, I^{v})), 
\end{equation} 
where $ \lambda_{\text{LPIPS}} $ is set to 0.05.

\noindent
\textbf{Semantic Loss ($\mathcal{L}_{\text{sem}}$).} 
Instead of purely distilling 2D LSeg features as in \cite{lsm,uni3r}, we leverage the proposed mix-training curriculum (Eq. \ref{eq:total_sem_loss}) to enhance semantic coherence and cross-scene generalization.

\noindent
\textbf{Geometric Loss ($\mathcal{L}_{\text{geo}}$).} 
Following \cite{uni3r}, we utilize a frozen geometric teacher (VGGT) \cite{vggt} to guide point maps $\hat{P}^{v}$ supervision: 
\begin{equation}
\mathcal{L}_{\text{geo}} = \sum_{v=1}^{V}\mathcal{L}_{\text{Chamfer}}\!\left(\hat{P}^{v}, \tilde{P}^{v}, \tilde{C}^{v}\right),
\end{equation}
where $\mathcal{L}_{\text{Chamfer}}$ is the Chamfer-style alignment loss used in \cite{uni3r}, computed between the predicted point map $\hat{P}^{v}$ and the teacher-provided point map $\tilde{P}^{v}$, and weighted by the teacher’s confidence map $\tilde{C}^{v}$.

\section{Experiments}  \label{sec:experiment}
\subsection{Experimental Setup}
\textbf{Datasets.} 
Following \cite{lsm,spatialsplat,uni3r}, we train our method on a combined dataset of ScanNet \cite{scannet} and ScanNet++ \cite{scannetpp}, covering 1,565 indoor scenes. 
For evaluation, 40 unseen scenes from ScanNet are used to assess the model’s performance on novel view synthesis, depth prediction, and open-vocabulary 3D segmentation. 
We further test on Replica \cite{replica} to assess cross-dataset generalization.

\noindent
\textbf{Implementation Details.} 
We use DINOv2 \cite{dinov2} as the image encoder with a patch size of 16, and a 24-layer cross-view Transformer. The encoder and decoder are initialized from the pretrained VGGT weights \cite{vggt}, while other parameters are randomly initialized.
All methods are evaluated at 256$\times$256 resolution for fair comparison. 
The framework supports up to $N$ = 16 input views per scene.
Training runs for 100 epochs on 8 A100 GPUs, taking $\sim$20 hours for the 2-view setting with batch size 32. 
SAM2 is used online during training, adding $\sim$4 hours. 
We also evaluate generalization under 8-view and 16-view inputs.

\noindent
\textbf{Evaluation Metrics.} 
We evaluate novel view synthesis using PSNR, SSIM \cite{ssim}, and LPIPS \cite{lpips}.
Segmentation performance is assessed using the mean Intersection-over-Union (mIoU) and mean pixel accuracy (mAcc). 
For depth prediction, we report standard metrics: Absolute Relative Error (Rel) and Root Mean Square Error (RMSE).

\noindent
\textbf{Baselines.} 
We compare our approach with state-of-the-art (SOTA) methods for generalizable semantic 3D reconstruction: LSM \cite{lsm}, SpatialSplat \cite{spatialsplat}, and Uni3R \cite{uni3r}. 
In addition, we incorporate the following baselines: 1) Per-scene optimization methods for semantic 3D reconstruction, including DFF \cite{DFF} and Feat3DGS \cite{feature3dgs}; and 2) the 2D open-vocabulary segmenter LSeg \cite{LSeg}, which is used for feature lifting in all compared semantic-aware methods.

\begin{table*}[t]
\renewcommand{\arraystretch}{1.1}
\centering 
\caption{
\textbf{Quantitative comparison on 3D tasks using 2-view inputs.} 
We report depth estimation quality, open-vocabulary segmentation accuracy, and novel view synthesis. 
Rel and RMSE values are scaled by 100 for readability. 
Results of Uni3R are obtained using its official GitHub implementation and released weights. 
Our approach achieves {\textbf{superior performance}} across various metrics.   
}  
\vspace{-3mm}
\label{table:SOTA_full_2views}
\begin{adjustbox} {width=.99\linewidth}
\begin{tabular}{ c | c | c | p{1.3cm}<{\centering} p{1.3cm}<{\centering} | p{1.4cm}<{\centering} p{1.4cm}<{\centering} | p{1.3cm}<{\centering} p{1.3cm}<{\centering}  p{1.3cm}<{\centering} p{1.3cm}<{\centering} }
\Xhline{3\arrayrulewidth}
\multirow{2}{*}{Type} & \multirow{2}{*}{Method} & \multirow{2}{*}{Recon. Time} & \multicolumn{2}{c|}{Depth Estimation} & \multicolumn{2}{c|}{Text-Driven Seg.} & \multicolumn{3}{c}{Novel View Synthesis}\\ 
 &  &  & Rel $\downarrow$ & RMSE $\downarrow$ & mIoU $\uparrow$ & mAcc $\uparrow$ & PSNR $\uparrow$ & SSIM $\uparrow$ & LPIPS $\downarrow$  \\  
\hline 
\multirow{3}{*}{\textit{Per-Scene}} 
&      LSeg        \cite{LSeg}   &  N/A   & N/A     & N/A     & 0.4819 & 0.7927 & N/A   & N/A    & N/A    \\
& DFF    \cite{DFF}              &  1m21s   & N/A     & N/A     & 0.4037 & 0.6755 & 19.86 & 0.6650 & 0.3629 \\
& Feat3DGS \cite{feature3dgs}    &  18m21s & N/A     & N/A     & 0.4223 & 0.7174 & 24.49 & 0.8132 & 0.2293 \\
\hline 
\multirow{4}{*}{\textit{Feed-Forward}} 
&  LSM \cite{lsm}      &   0.108s          & 4.74    & 18.90   & 0.5078 & 0.7686 & 24.39 & 0.8072 & 0.2506 \\
& SpatialSplat \cite{spatialsplat}  & 0.071s  & N/A     & N/A     & 0.5587 & 0.7924 & 25.46 & 0.8045 & 0.2046 \\
& Uni3R \cite{uni3r}          &  0.162s  & 4.85    & 15.37   & 0.5474 & 0.8198 & 25.37 & 0.8671 & 0.1412 \\
&  \cellcolor{lightblue}Ours  &  \cellcolor{lightblue}0.162s
& \cellcolor{lightblue}\textbf{3.84}   & \cellcolor{lightblue}\textbf{13.13} 
& \cellcolor{lightblue}\textbf{0.5677} & \cellcolor{lightblue}\textbf{0.8345} 
& \cellcolor{lightblue}\textbf{25.58}  & \cellcolor{lightblue}\textbf{0.8723} & \cellcolor{lightblue}{\textbf{0.1397}}  \\  
\Xhline{3\arrayrulewidth}
\end{tabular}
\end{adjustbox}
\end{table*}

\begin{table*}[t]
\renewcommand{\arraystretch}{1.1}
\centering 
\caption{\textbf{Quantitative comparison on 3D tasks with 8-view and 16-view inputs}. Our approach achieves {\textbf{superior performance}}. 
}  
\vspace{-3mm}
\label{table:SOTA_full_Multiviews}
\begin{adjustbox} {width=\linewidth}
\begin{tabular}{ c | p{0.8cm}<{\centering} c c c c c c | p{0.8cm}<{\centering} c c c c c c c c  }
\Xhline{3\arrayrulewidth}
\multirow{2}{*}{Method} & \multicolumn{7}{c|}{8 View Inputs} & \multicolumn{7}{c}{16 View Inputs}\\ 
& Rel  & RMSE  & mIoU  & mAcc  & PSNR  & SSIM & LPIPS  
& Rel  & RMSE  & mIoU  & mAcc  & PSNR  & SSIM & LPIPS \\ 
\hline 
Uni3R \cite{uni3r}          & 4.46    & 15.04   & 0.521 & 0.807 & 24.122 & 0.830 & 0.251 & 4.81 & 14.57  & 0.522 & 0.795 & 23.070 & 0.807 & 0.299 \\
\rowcolor{lightblue} Ours  & \textbf{3.73} & \textbf{12.94} & \textbf{0.533} & \textbf{0.821} & \textbf{24.520} & \textbf{0.843} & \textbf{0.162} & \textbf{4.08} & \textbf{13.41} & \textbf{0.552} & \textbf{0.832} & \textbf{23.668} & \textbf{0.823} & \textbf{0.184}  \\ 
\Xhline{3\arrayrulewidth}
\end{tabular}
\end{adjustbox}
\end{table*}

\subsection{Comparison with State-of-the-art Methods} \label{section:comparison_SOTA}  
\begin{figure}[t]
\centering

\begin{minipage}[t]{0.49\textwidth}
  \vspace{0pt}\centering
  \includegraphics[width=.98\linewidth]{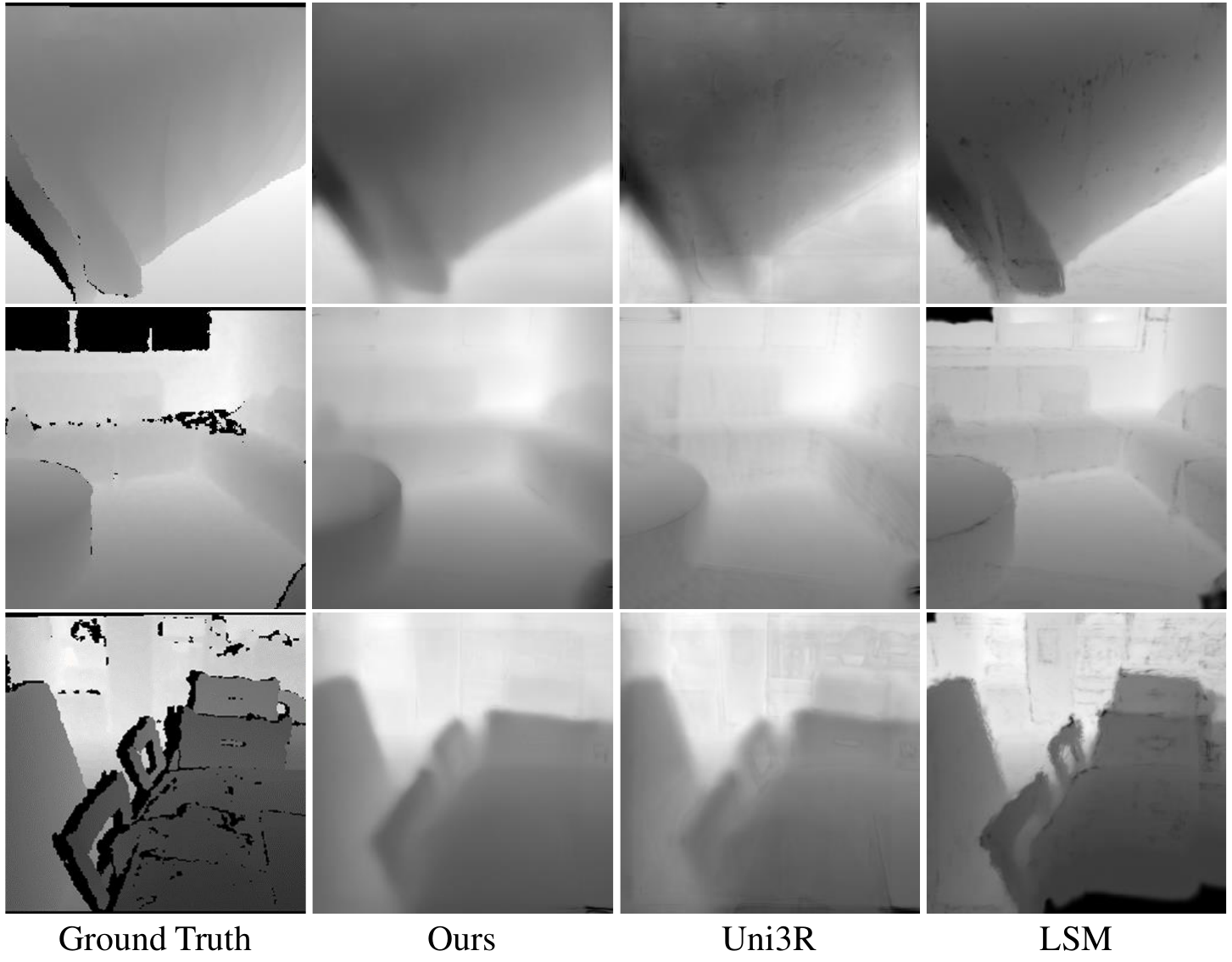}
  \vspace{-3.5mm}
  \captionof{figure}{\textbf{Depth estimation results on novel views.}}
  \label{fig:vis_comparison_depth}
\end{minipage}\hfill
\begin{minipage}[t]{0.49\textwidth}
  \vspace{0pt}\centering
  \includegraphics[width=.98\linewidth]{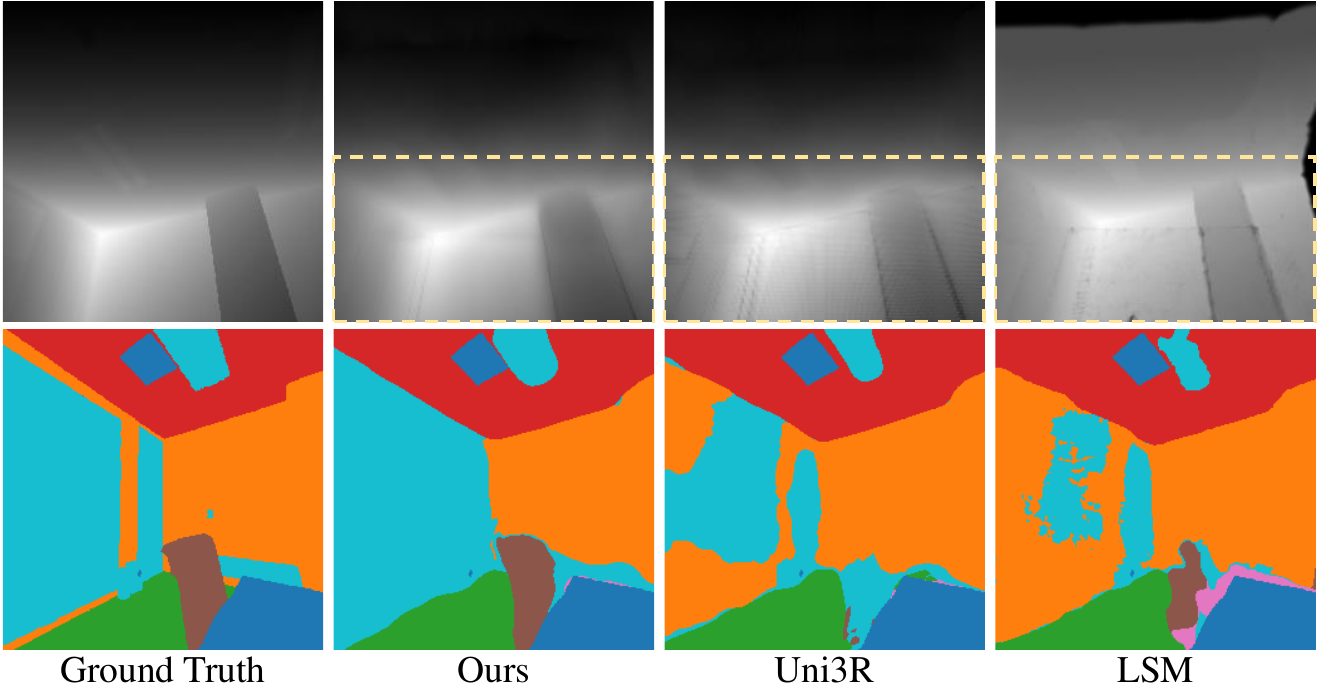}
  \vspace{-1mm}
  \captionof{figure}{\textbf{
  Qualitative comparison of cross-dataset generalization,} including depth estimation (1$^{st}$ row) and open-vocabulary segmentation (2$^{nd}$ row). Please zoom in for better visualization. 
  }
  \label{fig:vis_comparsion_replica}
\end{minipage}
  \vspace{-5.5mm}
\end{figure}

\textbf{Results of Depth Estimation.} 
In Table \ref{table:SOTA_full_2views}, UniSem achieves a significant 19.0\% reduction in Rel (from 4.74 to 3.84, scaled by 100) over prior methods. 
Fig. \ref{fig:vis_comparison_depth} shows that UniSem reconstructs more detailed and structurally coherent geometry across diverse and challenging scenes. 
These results exhibit its ability to learn effective Gaussian representations, enabling superior depth quality.

\begin{figure*}[!t]
\centering
\includegraphics[width=\linewidth]{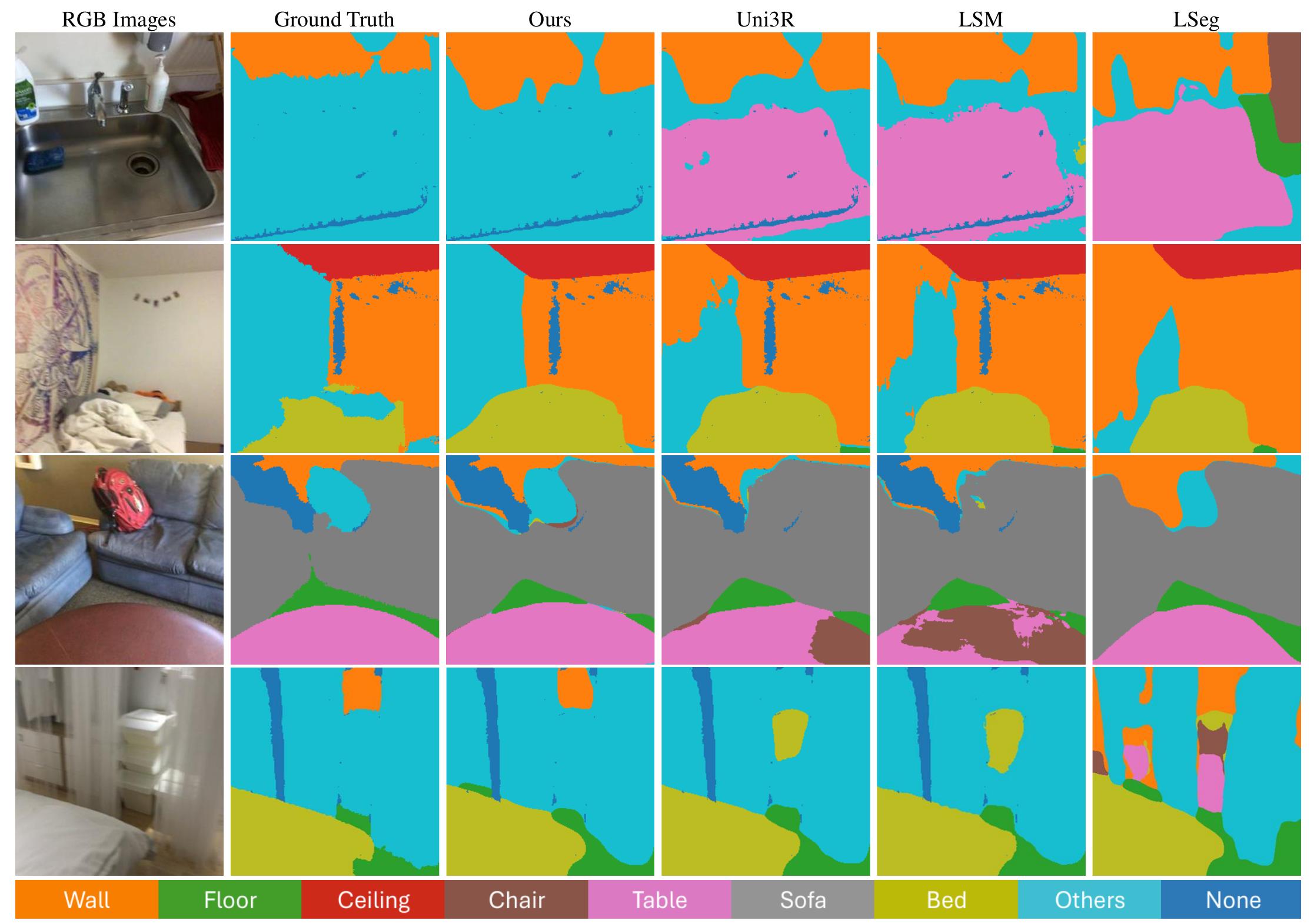}
\vspace{-7mm}
\caption{
\textbf{Open-vocabulary segmentation comparison on novel views.} 
Our approach yields more accurate and complete results. 
}
\label{fig:vis_comparison_seg}
  \vspace{-3mm}
\end{figure*}

\noindent
\textbf{Results of Open-Vocabulary 3D Segmentation.}
Following LSM and Uni3R, we remap the semantic categories in ScanNet into eight common classes: Wall, Floor, Ceiling, Chair, Table, Bed, Sofa, and Others. 
As shown in Table \ref{table:SOTA_full_2views}, UniSem surpasses all baselines, even outperforming LSeg, which provides semantic feature supervision. This indicates that our model learns stronger 3D-consistent semantics than what 2D distillation alone provides. 
Fig. \ref{fig:vis_comparison_seg} further confirms that UniSem yields more complete and coherent segmentation. 
This stems from the Mix-training Curriculum, which injects 3D semantic priors and suppresses noisy 2D inconsistencies, thus enabling more accurate object segmentation and improved cross-scene generalization.

\noindent
\textbf{Results of Novel View Synthesis.} 
In Table \ref{table:SOTA_full_2views}, per-scene optimization methods (\eg, Feat3DGS) require known poses and lengthy tuning, yet yield only comparable fidelity.  
In contrast, UniSem performs pose-free reconstruction in under one second while attains competitive or better rendering quality. 
Moreover, compared with a strong feed-forward method specialized for view synthesis (AnySplat), UniSem matches PSNR/SSIM and improves LPIPS (Table~\ref{table:sota_nvs}).

\noindent
\textbf{Scalability to Different View Inputs.} 
Unlike LSM and SpatialSplat, which are restricted to 2-view inputs, UniSem supports multi-view input, such as 8-view and 16-view settings as in Uni3R. 
Table \ref{table:SOTA_full_Multiviews} shows that UniSem consistently outperforms Uni3R across all view configurations. 
With 16 views, Rel decreases by 15.2\%, while mAcc rises by 3.7\%. 
These results highlight the robustness of EGD and MTC in handling multi-view inputs and wide baselines, achieving high-fidelity and semantically coherent 3D reconstructions.

\begin{table}[t]
\centering

\begin{minipage}[t]{0.38\textwidth}
  \vspace{0pt}
  \centering
  \caption{\textbf{Novel view synthesis comparison under 2-view inputs. }} 
  \label{table:sota_nvs}
  \vspace{-3mm}
    \begin{adjustbox} {width=.98\linewidth}
    \begin{tabular}{ c | p{1.4cm}<{\centering} p{1.4cm}<{\centering} p{1.4cm}<{\centering}  }
    \Xhline{3\arrayrulewidth}
    Setting  & PSNR $\uparrow$ & SSIM $\uparrow$ & LPIPS $\downarrow$ \\   
    \hline 
    AnySplat \cite{anysplat}     & 25.559 & 0.8707 & 0.241     \\ 
    Ours                         & 25.579 & 0.8723 & 0.140     \\ 
    \Xhline{3\arrayrulewidth}
    \end{tabular}
    \end{adjustbox}
  
\end{minipage}\hfill
\begin{minipage}[t]{0.58\textwidth}
  \vspace{0pt}
  \centering
  \caption{\textbf{Comparison on Replica with 2-view inputs.} Our approach achieves superior cross-dataset results.} 
  \label{table:SOTA_full_Replica}
  \vspace{-3mm}
  \begin{adjustbox} {width=0.98\linewidth}
    \begin{tabular}{ c | p{1.3cm}<{\centering} p{1.3cm}<{\centering} p{1.3cm}<{\centering} p{1.3cm}<{\centering} p{1.3cm}<{\centering} p{1.3cm}<{\centering}  p{1.3cm}<{\centering} }
    \Xhline{3\arrayrulewidth}
    Method  & Rel $\downarrow$  & RMSE $\downarrow$  & mIoU $\uparrow$  & mAcc $\uparrow$  & PSNR $\uparrow$  & SSIM $\uparrow$  & LPIPS  $\downarrow$  \\ 
    \hline 
    LSM \cite{lsm}              & 13.71         & 32.56          & 0.3596          & 0.7054          & 13.219         & 0.585           & 0.346     \\ 
    Uni3R \cite{uni3r}          & 7.16          & 20.20          & 0.4252          & 0.7616          & 23.888         & 0.829           & 0.115          \\
    \rowcolor{lightblue} Ours   & \textbf{6.52} & \textbf{18.38} & \textbf{0.4784} & \textbf{0.7966} & \textbf{24.448} & \textbf{0.849}  & \textbf{0.111}  \\ 
    \Xhline{3\arrayrulewidth}
    \end{tabular}
  \end{adjustbox}
  
\end{minipage}
\vspace{-3mm}
\end{table}

\noindent
\textbf{Cross-Dataset Generalization.}
We evaluate UniSem on the Replica without any fine-tuning.
In Table \ref{table:SOTA_full_Replica}, UniSem significantly outperforms prior methods in depth estimation, semantic segmentation, and novel view synthesis. 
Fig. \ref{fig:vis_comparsion_replica} also highlights improved depth and more accurate semantics.
These confirm the strong cross-domain generalization capability enabled by our EGD and MTC.

\begin{table}[t]
\centering

\begin{minipage}[t]{0.48\textwidth}
  \vspace{0pt}
  \centering
  \caption{\textbf{Main ablation study} under 2-view input setting.}
  \label{table:ablation_main}
  \vspace{-3mm}
  \begin{adjustbox} {width=0.98\linewidth}
  \begin{tabular}{ c | p{0.9cm}<{\centering} p{1.0cm}<{\centering} p{1.0cm}<{\centering} p{1.0cm}<{\centering} p{1.0cm}<{\centering} p{1.0cm}<{\centering} }
    \Xhline{3\arrayrulewidth}
    Setting & Rel & RMSE & mIoU & mAcc & PSNR & SSIM \\
    \midrule[0.1pt]
    \rowcolor{lightblue} Ours (Full)
    & \textbf{3.84} & \textbf{13.13} & \textbf{0.568} & \textbf{0.835} & \textbf{25.58} & \textbf{0.872} \\
    \midrule[0.1pt]
    w/o EGD          & 4.47 & 14.30 & 0.557 & 0.831 & 25.29 & 0.869 \\
    w/o Error-aware  & 3.94 & 13.31 & 0.561 & 0.832 & 25.35 & 0.870 \\
    w/o Cosine-cycle & 3.95 & 13.26 & 0.555 & 0.830 & 25.36 & 0.868 \\
    \midrule[0.1pt]
    w/o MTC          & 4.09 & 13.68 & 0.555 & 0.819 & 25.36 & 0.869 \\
    w/o $\mathcal{L}^{\text{3D}}_p$ & 3.92 & 13.29 & 0.563 & 0.834 & 25.46 & 0.870 \\
    w/o Geo-aware    & 3.87 & 13.26 & 0.565 & 0.835 & 25.54 & 0.871 \\
    \Xhline{3\arrayrulewidth}
  \end{tabular}
  \end{adjustbox}
  
\end{minipage}\hfill
\begin{minipage}[t]{0.48\textwidth}
  \vspace{0pt}
  \centering
  \caption{\textbf{Comparison with dropout-based strategies in 3DGS} (DropGaussian~\cite{dropgaussian}, DropoutGS~\cite{dropoutgs}, and CAGS~\cite{chen2025quantifying}).}
  \label{table:ablation_egd}
  \vspace{-3mm}
    \begin{adjustbox} {width=.94\linewidth}
    \begin{tabular}{ c | p{0.9cm}<{\centering} p{1.0cm}<{\centering} p{1.0cm}<{\centering} p{1.0cm}<{\centering} p{1.0cm}<{\centering} p{1.0cm}<{\centering}  p{1.0cm}<{\centering}  }
    \Xhline{3\arrayrulewidth}
    Setting & Rel  & RMSE  & mIoU  & mAcc  & PSNR  & SSIM    \\   
    \midrule[0.1pt]
    \rowcolor{lightblue} Ours  
    & \textbf{3.84} & \textbf{13.13} 
    & \textbf{0.568} & \textbf{0.835} 
    & \textbf{25.58} & \textbf{0.872}    \\ 
    \midrule[0.1pt]
    w \cite{dropgaussian}           & 12.30         & 33.50          & 0.494          & 0.801          & 18.68          & 0.734          \\ 
    w \cite{dropoutgs}              & 10.07         & 31.35          & 0.319          & 0.674          & 19.87          & 0.722        \\ 
    w \cite{chen2025quantifying}    & 4.21          & 13.99          & 0.557          & 0.819          & 25.21          & 0.865      \\ 
    \Xhline{3\arrayrulewidth}
    \end{tabular}
    \end{adjustbox}

\end{minipage}
\vspace{-3mm}
\end{table}

\begin{figure}[t]
\centering

\begin{minipage}[t]{0.49\textwidth}
  \vspace{0pt}\centering
  \includegraphics[width=.98\linewidth]{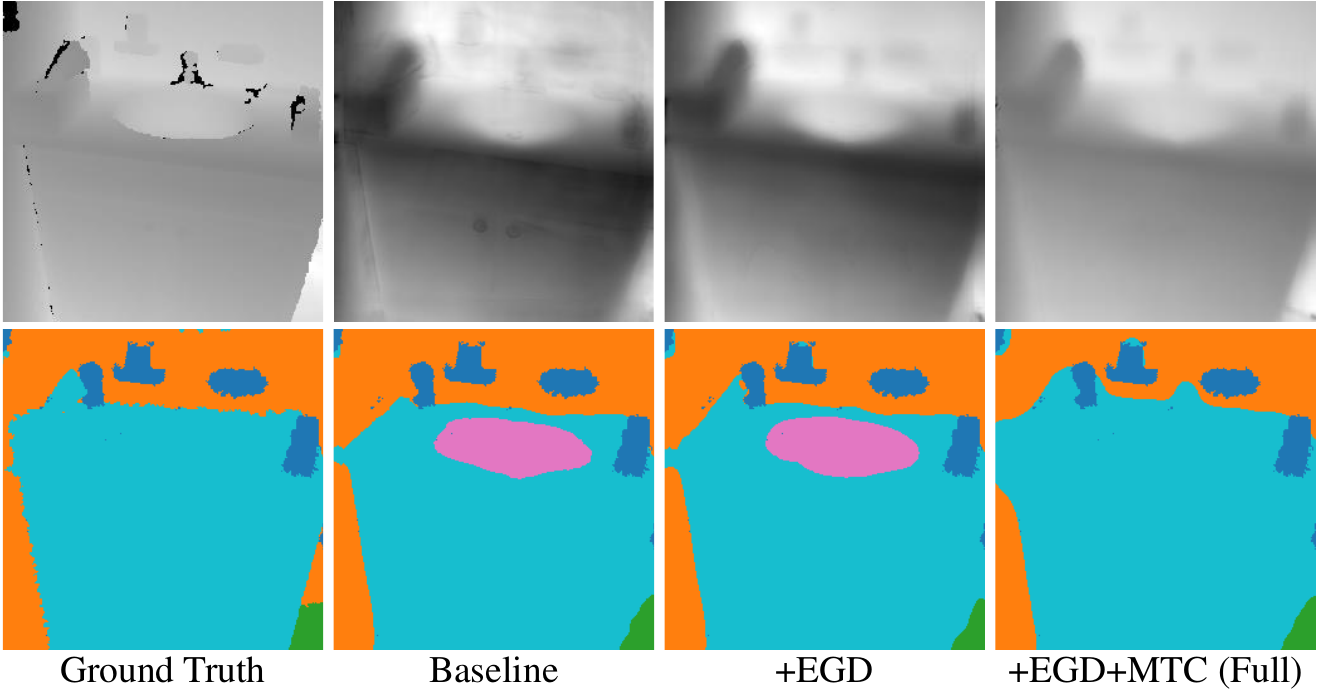}
  \vspace{-1mm}
  \captionof{figure}{\textbf{Visual ablation of key components.} EGD reduces depth noise, and MTC improves segmentation accuracy. }  
  \label{fig:vis_ablation}
\end{minipage}\hfill
\begin{minipage}[t]{0.49\textwidth}
  \vspace{0pt}\centering
  \includegraphics[width=.98\linewidth]{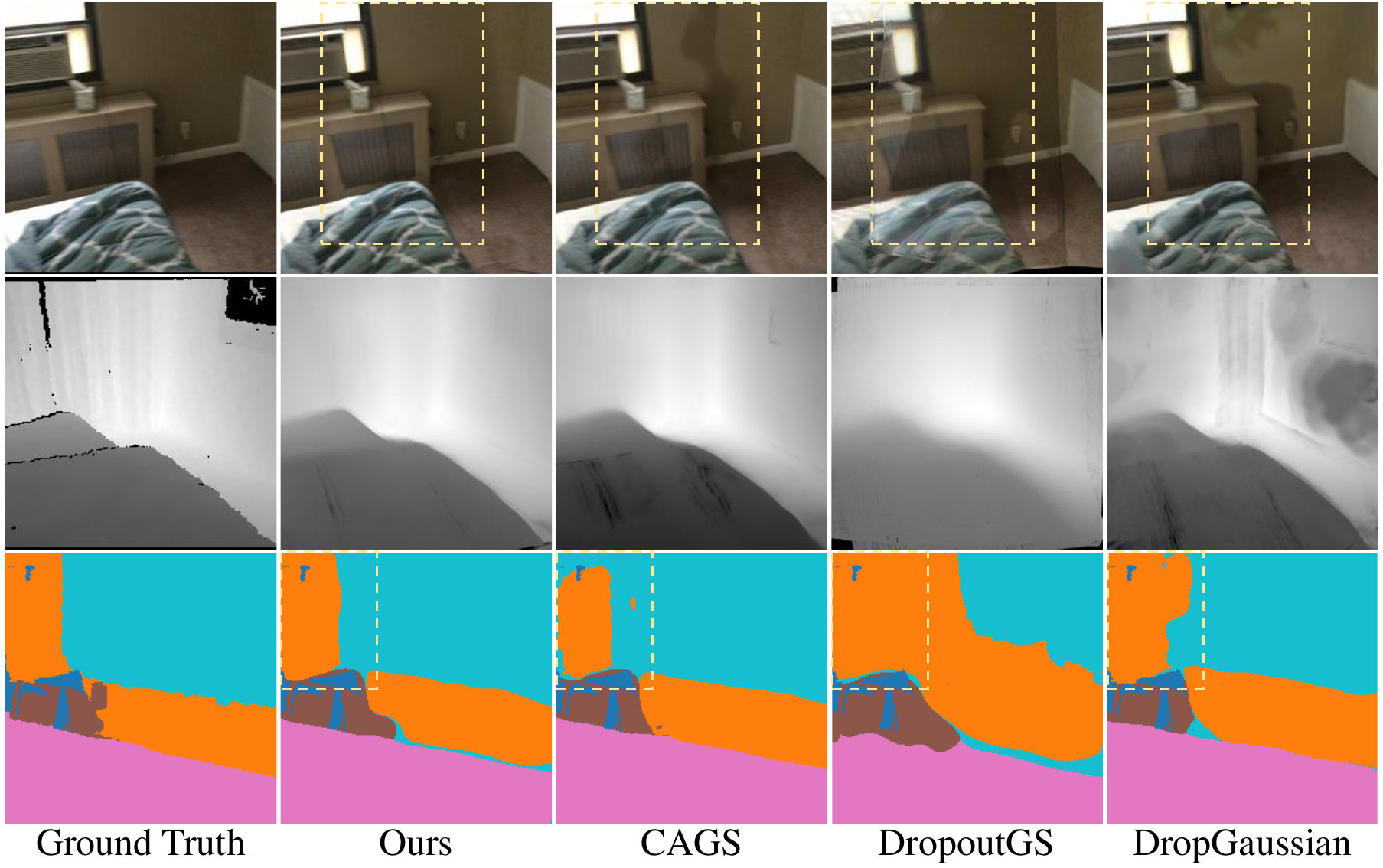}
  \vspace{-3mm}
  \captionof{figure}{\textbf{Visual comparison of our EGD} with different settings.  }
  \label{fig:vis_comparison_EGD}
\end{minipage}
\vspace{-6mm}
\end{figure}

\subsection{Ablation Studies} \label{section:ablation_study}  
\textbf{Effectiveness of EGD.}  
In Table \ref{table:ablation_main}, removing Error-aware Gaussian Dropout (EGD) consistently degrades all metrics, with the largest drop in depth, indicating that EGD regularizes and stabilizes Gaussian optimization. 
Replacing error-aware selection with naive random dropout (\textit{w/o Error-aware}) harms rendering quality, since error-agnostic removal may discard critical Gaussians and disrupt rendering. 
Using a fixed rather than a cosine-cycled dropout ratio (\textit{w/o Cosine-cycle}) degrades performance, confirming the scheduling strategy's role in stable optimization.

\noindent
\textbf{Effectiveness of MTC.}  
Table \ref{table:ablation_main} shows that Mix-training Curriculum (MTC) boosts segmentation, confirming its benefit for 3D semantic generalization. 
Beyond semantics, its correspondence alignment also benefits depth and appearance rendering under sparse views.  
Both the 3D correspondence loss (\textit{w/o $\mathcal{L}^{\text{3D}}_p$}) and its geometry-aware weighting (\textit{w/o Geo-aware}) contribute to consistent improvements, validating their roles in reliable feature alignment.
Fig. \ref{fig:vis_ablation}  further show reduced depth noise with EGD and more coherent segmentation with MTC.

\noindent
\textbf{Variants of EGD.}  
Table \ref{table:ablation_egd} compares EGD with related dropout-based strategies in 3DGS.   
DropGaussian \cite{dropgaussian} indiscriminately zeros out opacities, causing depth and rendering instability. 
DropoutGS \cite{dropoutgs} enforces invariance between full-field and Gaussian-dropped renderings, assuming that removing redundant Gaussians should not alter the representation. Yet, this overly strict constraint destabilizes optimization and hurts accuracy.  
CAGS \cite{chen2025quantifying} uses fixed-ratio random dropout. However, its purely random removal and the lack of a ratio regulation mechanism introduce noisy, unstable gradients, which perturb the radiance field and harm geometry and rendering.  
In contrast, EGD performs selective, error-aware suppression with a cosine-cycle schedule, gradually regularizing redundant Gaussians while retaining critical ones.
This yields more stable feed-forward Gaussian optimization and consistently better results (\ref{fig:vis_comparison_EGD}).

\noindent
\textbf{Ablation on VGGT Geometry Loss.}
Table~\ref{table:more_ablation_vgg_geo} shows that under \emph{w/o Geo}, our 
\begin{wraptable}{r}{0.48\textwidth}
\vspace{-11mm}
\caption{\textbf{EGD beyond VGGT geometric supervision.} 
Even under \emph{w/o Geo}, our method remains superior.
}
\label{table:more_ablation_vgg_geo}
\vspace{-1mm}
\centering
\begin{adjustbox}{width=\linewidth}
\begin{tabular}{ c | p{0.85cm}<{\centering} p{0.95cm}<{\centering} p{0.95cm}<{\centering} p{0.95cm}<{\centering} p{0.95cm}<{\centering} p{0.95cm}<{\centering} p{0.95cm}<{\centering} }
\Xhline{3\arrayrulewidth}
Method & Rel & RMSE & mIoU & mAcc & PSNR & SSIM \\
\midrule[0.1pt]
Uni3R w/o Geo & 7.73 & 22.21 & 0.543 & 0.820 & 23.47 & 0.798 \\
Ours  w/o Geo & 5.89 & 18.73 & 0.552 & 0.834 & 25.51 & 0.868 \\
\Xhline{3\arrayrulewidth}
\end{tabular}
\end{adjustbox}
\vspace{-8.8mm}
\end{wraptable}
method still surpasses Uni3R on depth and rendering. 
This suggests that the improvement is not solely attributable to VGGT  geometric supervision, and EGD provides effective regularization that stabilizes Gaussian optimization.

\noindent
\textbf{Effect of the Cosine-cycle Schedule in EGD.}
Table~\ref{table:more_ablation_egd} compares cosine-cycle with fixed and step schedules.
A fixed ratio (always $\eta_{\max}$) is overly aggressive early on, destabilizing Gaussian optimization and degrading semantics/rendering, while a step schedule ($\eta_{\min}\!\rightarrow\!\eta_{\max}$) improves early stability but becomes too strong later, slightly hurting perceptual quality (LPIPS).
Cosine-cycle provides a balanced progression: mild dropout at the beginning/end for stable geometry and detail preservation, and stronger dropout mid-training for redundancy reduction, yielding superior overall results.

\noindent
\textbf{Variants of MTC.}
In Table~\ref{table:more_ablation_mtc}, replacing max-error prompting with random points reduces mIoU/mAcc, since random prompts may fall on semantically uninformative regions and provide weaker correction on challenging areas. Max-error prompting instead targets high-uncertainty locations, improving cross-view correspondence and semantic propagation. 
Earlier activation (e.g., $\tau=80$ rather than\ $\tau=90$) achieves comparable accuracy but increases overhead for correspondence construction (via SAM2), so we set $\tau=90$ for an efficient trade-off.

\begin{table}[t]
\centering

\begin{minipage}[t]{0.48\textwidth}
  \vspace{0pt}
  \centering
\caption{Ablation results of EGD with different settings. 
}  
\label{table:more_ablation_egd}
\vspace{-3mm}
\begin{adjustbox} {width=.98\linewidth}
\begin{tabular}{ c | p{0.9cm}<{\centering} p{1.0cm}<{\centering} p{1.0cm}<{\centering} p{1.0cm}<{\centering} p{1.0cm}<{\centering} p{1.0cm}<{\centering}  p{1.0cm}<{\centering}  }
\Xhline{3\arrayrulewidth}
Method & Rel  & RMSE  & mIoU  & mAcc  & PSNR  & LPIPS    \\   
\midrule[0.1pt]
\rowcolor{lightblue}  Cosine-cycle  
& \textbf{3.84} & \textbf{13.13} 
& \textbf{0.568} & \textbf{0.835} 
& \textbf{25.58} & \textbf{0.1397}    \\ 
\midrule[0.1pt]
Fix Schedule     & 3.95    & 13.26   & 0.555 & 0.830  & 25.36   & 0.1455    \\
Step Schedule    & 3.89    & 13.24   & 0.565 & 0.834  & 25.50   & 0.1417    \\ 
\Xhline{3\arrayrulewidth}
\end{tabular}
\end{adjustbox}

\end{minipage}\hfill
\begin{minipage}[t]{0.48\textwidth}
  \vspace{0pt}
  \centering
    \caption{Ablation results of MTC with different settings. 
    }  
    \label{table:more_ablation_mtc}
  \vspace{-3mm}
    \begin{adjustbox} {width=.98\linewidth}
    \begin{tabular}{ c | p{0.9cm}<{\centering} p{1.0cm}<{\centering} p{1.0cm}<{\centering} p{1.0cm}<{\centering} p{1.0cm}<{\centering} p{1.0cm}<{\centering}  p{1.0cm}<{\centering}  }
    \Xhline{3\arrayrulewidth}
    Method & Rel  & RMSE  & mIoU  & mAcc  & PSNR  & SSIM    \\   
    \midrule[0.1pt]
    \rowcolor{lightblue}  Ours       
    & \textbf{3.84} & \textbf{13.13} 
    & \textbf{0.568} & \textbf{0.835} 
    & \textbf{25.58} & \textbf{0.872}    \\ 
    \midrule[0.1pt]
    Random Prompt.     & 3.85    & 13.13   & 0.561  & 0.830  & 25.56  & 0.872    \\ 
    $\tau=80$          & 3.84    & 13.22   & 0.569  & 0.835  & 25.59  & 0.873    \\ 
    \Xhline{3\arrayrulewidth}
    \end{tabular}
    \end{adjustbox}

\end{minipage}
\vspace{-3mm}
\end{table}

\subsection{Limitation Analysis} \label{Analysis_limitations} 
UniSem is trained and evaluated primarily on indoor benchmarks, which are smaller and less diverse than datasets used by large-scale models (e.g., MASt3R~\cite{mast3r}, VGGT~\cite{vggt}). Thus, our current claims focus on indoor generalization under sparse, unposed inputs, and transfer to in-the-wild domains (outdoor, extreme lighting, dynamic scenes) will be explored in the future. 
Moreover, while SAM2-based region masks effectively provide cross-view correspondence, they may introduce segmentation artifacts in highly cluttered scenes. We hope to address this in the future by exploring mask-free semantic priors.

\section{Conclusion} \label{sec:conclusion}
This paper presents UniSem, a feed-forward framework that addresses the challenges of generalizable semantic-aware 3D reconstruction. 
By coupling Error-aware Gaussian Dropout (EGD), which suppresses redundancy-induced gradient interference, with the Mix-training Curriculum (MTC), which incorporates emergent 3D semantic cues via prototype-based alignment, UniSem jointly advances depth estimation and open-vocabulary 3D semantics. 
Experiments across varying numbers of input views and multiple benchmarks demonstrate consistent improvements in depth prediction, open-vocabulary 3D segmentation, novel-view synthesis, and strong cross-scene generalization.

\clearpage 

\bibliographystyle{splncs04}
\bibliography{main}
\end{document}